\documentclass[letterpaper]{article} 
\usepackage{aaai25}  
\usepackage{times}  
\usepackage{helvet}  
\usepackage{courier}  
\usepackage[hyphens]{url}  
\usepackage{graphicx} 
\usepackage{natbib}  
\usepackage{caption} 
\usepackage{algorithm}
\usepackage{algorithmic}

\usepackage{booktabs}%
\usepackage{amsmath,amssymb,amsfonts,amsthm}%
\usepackage{todonotes}
\usepackage{pifont}

\DeclareMathOperator*{\argmax}{arg\,max}
\DeclareMathOperator*{\argmin}{arg\,min}

\newtheorem{definition}{Definition}

\usepackage{newfloat}
\usepackage{listings}
\DeclareCaptionStyle{ruled}{labelfont=normalfont,labelsep=colon,strut=off} 
\floatstyle{ruled}
\newfloat{listing}{tb}{lst}{}
\floatname{listing}{Listing}
\title{Generating Synthetic Fair Syntax-agnostic Data by Learning and Distilling Fair Representation}
\author{
    Md Fahim Sikder\textsuperscript{\rm 1},
    Resmi Ramachandranpillai\textsuperscript{\rm 2},
    Daniel de Leng\textsuperscript{\rm 1}, and
    Fredrik Heintz\textsuperscript{\rm 1}
}
\affiliations{
    \textsuperscript{\rm 1}Department of Computer and Information Science (IDA), Link\"oping University, Sweden\\
	 \textsuperscript{\rm 2}Institute for Experiential AI, Northeastern University, USA\\
	 md.fahim.sikder@liu.se, r.ramachandranpillai@northeastern.edu, \\ \{daniel.de.leng, fredrik.heintz\}@liu.se
}

\usepackage{bibentry}

\begin{document}

\maketitle

\begin{abstract}


    Data Fairness is a crucial topic due to the recent wide usage of AI powered applications. Most of the real-world data is filled with human or machine biases and when those data are being used to train AI models, there is a chance that the model will reflect the bias in the training data. Existing bias-mitigating generative methods based on GANs, Diffusion models need in-processing fairness objectives and fail to consider computational overhead while choosing computationally-heavy architectures, which may lead to high computational demands, instability and poor optimization performance. To mitigate this issue, in this work, we present a fair data generation technique based on knowledge distillation, where we use a small architecture to distill the fair representation in the latent space.  The idea of fair latent space distillation enables more flexible and stable training of Fair Generative Models (FGMs). We first learn a syntax-agnostic (for any data type) fair representation of the data, followed by distillation in the latent space into a smaller model. After distillation, we use the distilled fair latent space to generate high-fidelity fair synthetic data. While distilling, we employ quality loss (for fair distillation) and utility loss (for data utility) to ensure that the fairness and data utility characteristics remain in the distilled latent space. Our approaches show a 5\%, 5\% and 10\% rise in performance in fairness, synthetic sample quality and data utility, respectively, than the state-of-the-art fair generative model.

\end{abstract}

%

\section{Introduction}
\label{sec:introduction}

With the availability of vast amounts of data, the use of Artificial Intelligence powered systems has increased exponentially. 
These systems are being used to ease our daily life tasks, e.g. language translation, health-care diagnosis. 
The enormous amount of available data is being used to train these AI-based systems. 
Unfortunately, these available datasets contain machine or human bias \cite{liu2022fair}. 
There is a chance that a trained model using these biased data will also give biased outcomes for certain demographics \cite{caton2024fairness}. 
For example, in the COMPAS case, the future risk assessment software gives high-risk results towards African-Americans due to false-positive outcomes \cite{compascase}. 
Recently, large language models have become increasingly popular.
However, these models were shown to discriminate against women and people with dark skin \cite{fang2024bias}. 
In another example of bias in decision-making, the Dutch Tax and Customs Administration in 2021 unlawfully leveraged residents' citizenship information to assess the risk of fraud in childcare benefits applications.
This resulted in selection bias in the decision-making process and caused a lot of agony to the affected people \cite{dutchscandal, weerts2023look}.
So, considering these incidents, the necessity of bias-free dataset or model that can mitigate bias is high.

Over the years, researchers have proposed various bias mitigation techniques for improving the model performance towards fairness and data utility. 
These techniques work either by pre-processing the data in a way before feeding it to a model that the dataset becomes fair (pre-processing), changing the model architecture in a way that the output of the model will satisfy fairness criteria (in-processing) or by changing the model outcome after training in a way that the outcome is fair towards demographics (post-processing) \cite{caton2024fairness}.

\begin{figure}[!t]
	\centering
	\includegraphics[width=0.65\columnwidth]{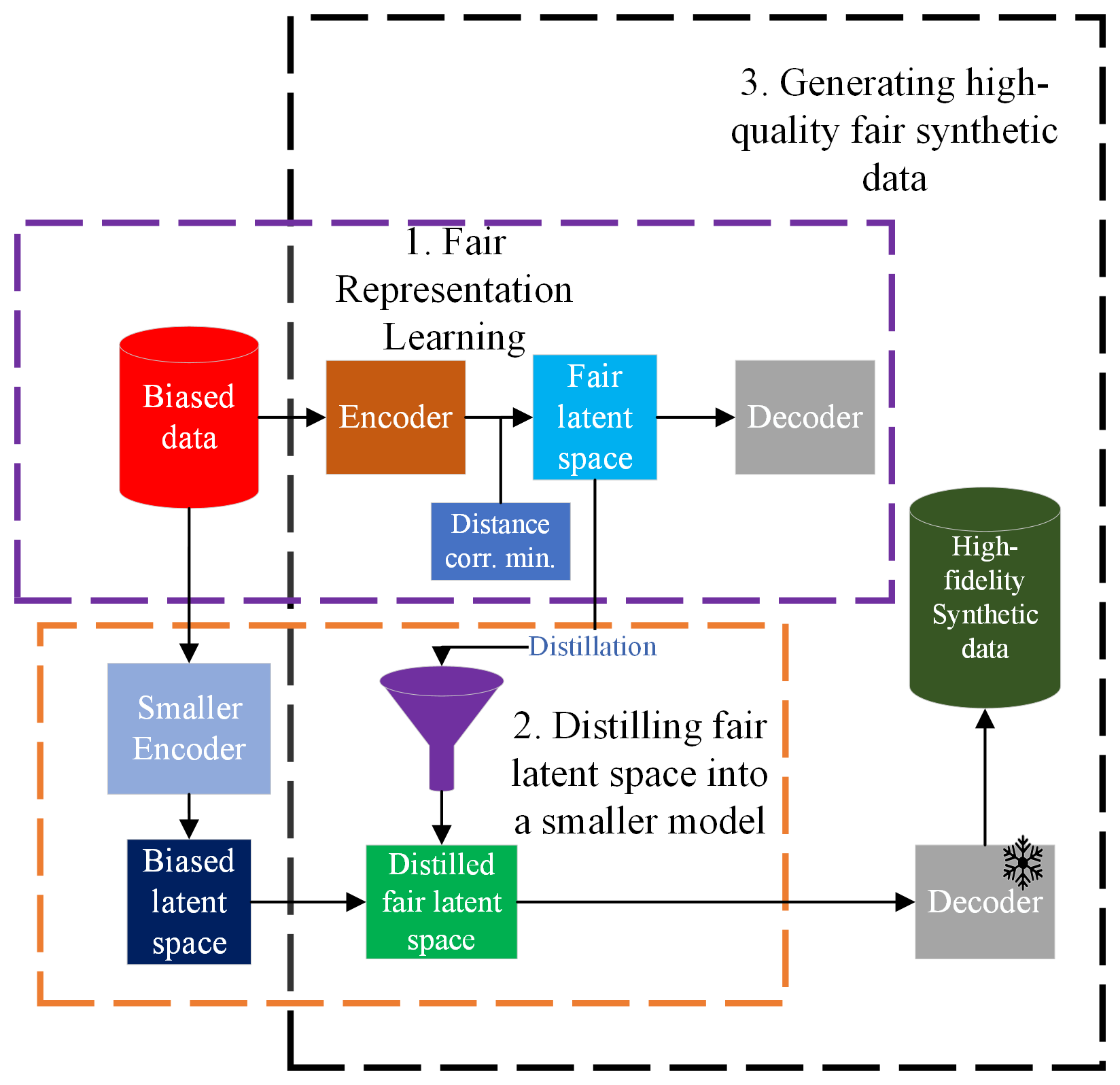}
	\caption{Our approach for fair latent distillation and synthetic sample generation}
	\label{fig:architecture}
\end{figure}

Representation learning refers to the process of learning latent features which can be useful for downstream tasks. Different approaches have been proposed to learn fair representation from biased datasets over time \cite{zemel2013learning, xu2021robust, maxalfr2023}. Besides creating fair representation, researchers also use generative models to generate fair synthetic samples \cite{rajabi2022tabfairgan, li2022fairgan, van2021decaf, ramachandranpillai2023fair, liu2022fair}. Though these fair representations and generative models achieve great performance on fairness, sometimes, some trade-offs exist between data fairness quality towards certain demographics and data utility in the downstream tasks \cite{dutta2020there}. On top of that, these models, especially the generative models, need extensive computational resources to generate fair synthetic samples.



Distillation is a process of transferring knowledge from a learned model to a smaller model. Using a smaller model helps to reduce the computational overhead in the training process and makes it possible to deploy the model in a smaller environment, e.g. edge devices. Most of the knowledge distillation work is based on the classification tasks \cite{dong2023reliant, zhu2024devil} and is heavily dependent on the data labels. For this reason, the distillation process is not directly applicable to representation learning, where the model learns the latent feature. 


From these motivations, in this work, we present a Syntax-agnostic (for any type of data) Fair Generative Model (FGM) that allows us to distill latent space and generates high-fidelity fair synthetic samples. In our model, first, we create a fair representation and then transfer the knowledge from these latent spaces to a smaller model using distillation. We develop a way to transfer the knowledge from trained fair model to a smaller model on latent space. After distillation, we use the distilled latent space to generate fair synthetic data. The disjunction of fairness optimization and synthetic sample generation helps to increase the stability and optimization for whole generative model process \cite{ramachandranpillai2023fair}. Also, as we use distillation on the latent space, we do not need to worry about the data types (syntax-agnostic), which allows us to reduce the computational overhead, and for using the smaller model, we also reduce the training time. During the distillation process, to keep the features of the fair representation in the distilled representation, we employ quality loss and utility loss to keep the data utility characteristics (details in section \ref{subsec:distillation-fair}). This allows us to perform better than the base fair model and with other fair generative models. 

Our Contributions in this work is as follows:

\begin{itemize}
    \item We present a novel Fair Generative Model (FGM) process to distill latent space and generate high-fidelity fair synthetic data (both tabular and image data).
    \item We show, our distilled model outperform the base fair model as well as state-of-the-art fair models in terms of fairness and data utility.
    \item By using, distillation in latent space, we reduce the computational overhead and training time.
\end{itemize}



\section{Related Works}
\label{sec:related-works}

In recent years, various techniques have been proposed to learn \emph{fair representation} from biased data. LFR \cite{zemel2013learning} presents representation learning approach as a optimization task and showed their work achieve fair performance in the downstreaming task. FairDisco \cite{liu2022fair} learns fair representation by minimizing the correlation between the sensitive attributes and non-sensitive attributes by using a variance correlation loss. Adversarial approach has been also used to learn fair representation \cite{gao2022fairneuron, madras2018learning, xu2021robust} where the adversary critiques probable unfair outcome.

\emph{Knowledge distillation} is a technique to transfer the knowledge of one model to another. This is usually done by comparing both the teacher (the model we want to distil) and student model's (the model where we want to transfer the knowledge) last layer output and using it to optimize the student model. Recently some work has been done in the field of distillation on data fairness. RELIANT \cite{dong2023reliant} and FairGKD \cite{zhu2024devil} uses Graph Neural Network and knowledge distillation to lean the fair representation of the data. However, one of the drawbacks of these approach is the data label is required for distillation, which makes it challenging to distill latent space. So, in this work, we present a novel approach to distil the latent space which eliminates the necessity of data label. 

Over the years, various \emph{(fair) generative models} have been proposed to generate synthetic fair data. Generative Adversarial Networks (GANs) based TabFairGAN \cite{rajabi2022tabfairgan} tries to generate fair data by adding fairness constraints during the training process, FLDGMs \cite{ramachandranpillai2023fair} produces synthetic fair data by generating latent space using GANs and Diffusion models. However, these models are expensive to train due to their architectural design, while our presented architecture is lightweight and takes less computational resources and less time to train.

\section{Background}
\label{sec:preliminaries}

In this section, we present the background information to follow the paper. First, we present what we mean by fairness, then the background of knowledge distillation, followed by information about synthetic data generation.

\subsection{Problem Formalization and Data Fairness}

For a given dataset $D:= (\{x_i, y_i, s_i\})_{i=1}^n$, here $(x_1, x_2, .. x_n)\in \mathcal{X}$ are non-sensitive attributes and i.i.d random variables, $s \in \mathcal{S}$ are the sensitive attributes, e.g: race, gender, etc. and $y \in \mathcal{Y}$ is the target variables. The goal here is to create a fair generative model $\mathcal{G}$ that generates synthetic dataset $D^{\prime}$ by creating and distilling fair representation. The fair representation can be obtained using a model (which can be an Encoder), $z = f_{\phi}(x,s)$, takes the input $(x, s) \in (\mathcal{X}, \mathcal{S})$ and produce a lower dimensional representation $z\in\mathcal{Z}$. Here, $\phi$ is the trainable parameters for the model $f$. The latent representation $z$ needs to be fair, so that for any downstreaming task, the output $\hat{y}\in \hat{Y}$ would be independent of the sensitive attribute $s$. In the rest of the paper, we use ``Fair Representation'' and ``Fair Latent Space'' alternatively.

Fairness is a very broad term, and it can mean different things in different discipline. In this paper, when we present fairness, we mean algorithmic fairness. A model will be algorithmic fair if it satisfies definition \ref{def:dp} and/or \ref{def:eo}:

\begin{definition}[Demographic Parity (DP), \citet{barocas2016big}]
\label{def:dp}
For a given dataset $D$, a model $H: X \rightarrow \hat{Y}$, $\hat{Y} = \{0, 1\}$ will satisfy demographic parity (DP), iff 
\begin{equation*}
    P[H(X) = 1 \mid \mathcal{S} = s] = P[H(X) = 1 \mid \mathcal{S} = s^{\prime}]
\end{equation*}
for all ${s,s^{\prime}}\in \mathcal{S}$.
\end{definition}

\begin{definition}[Equalized Odds (EO), \citet{barocas2016big}]
\label{def:eo}
For a given dataset $D$, a model $H: X \rightarrow \hat{Y}$, $\hat{Y} = \{0, 1\}$ will satisfy equalized odds (EO), iff
\begin{equation*}
    [H(X) = 1 \mid Y = 1, \mathcal{S} = s] = P[H(X) = 1 \mid Y = 1, \mathcal{S} = s^{\prime}]
\end{equation*}
for all ${s,s^{\prime}}\in \mathcal{S}$.
\end{definition}

\begin{definition}[Demographic Parity Ratio (DPR), \citet{weerts2023fairlearn}]
\label{def:dpr}
For a given dataset $D$, with different subgroup $s,s^{\prime} \in \mathcal{S}$, and a model $H: X \rightarrow \hat{Y}$, $\hat{Y} = \{0, 1\}$, the demographic parity ratio determines the selection rate of different subgroups. DPR score of $1$ means, each subgroup in the model receives same selection rate.

\begin{equation*}
\label{eq:dpr}
    DPR = \frac{P[H(X) = 1 \mid \mathcal{S} = s]}{P[H(X) = 1 \mid \mathcal{S} = s^{\prime}]}
\end{equation*}
\end{definition}

\begin{definition}[Equalized Odds Ratio (EOR), \citet{weerts2023fairlearn}]
\label{def:eor}
For a given dataset $D$, with different subgroup $s,s^{\prime} \in \mathcal{S}$, and a model $H: X \rightarrow \hat{Y}$, $\hat{Y} = \{0, 1\}$, the equlaized odds ratio determines the selection rate of different subgroups. EOR score of $1$ means, each subgroup receives the same true positive, false positive, true negative and false negative rates.
\begin{equation*}
\label{eq:eor}
    EOR = \frac{P[H(X) = 1 \mid Y = 1, \mathcal{S} = s]}{P[H(X) = 1 \mid Y = 1, \mathcal{S} = s^{\prime}]}
\end{equation*}
\end{definition}

\subsection{Knowledge Distillation}

Knowledge distillation is a transfer learning process where the knowledge of one trained model (teacher model) is being transferred to another smaller models (student model) \cite{hinton2015distilling}. Knowledge can be distilled from one model to another model by various manner, e.g. in the response-based knowledge distillation process, the knowledge can be transferred by mimicing the output layer of the teacher model \cite{hinton2015distilling, jung2021fair, dai2021general}. Usually, the \textit{Cross-entropy-loss (CE)} is used on the final layer output added with the \textit{Kl-divergence (KL)}. 

\begin{equation}
    \label{eq:distillation-origin}
    \mathcal{L}_{\text{distillation}} = \mathcal{L}_{\text{CE}}(y, \hat{y}) + \mathcal{L}_{\text{KL}}(p, q)
\end{equation}

Equation \ref{eq:distillation-origin} shows an example of distillation loss where $y, \hat{y}$ are the label and predicted label respectively and $p, q$ are two different distributions.  Sometimes in the loss function a parameter $T$ is used to control probability distribution from both the teacher and student model \cite{hinton2015distilling}. In the feature-based distillation, the knowledge can be transferred by mimicing the intermediate layer of the teacher model \cite{zhang2020improve}.

All of the aforementioned distillation, learns to mimic the teacher model based on the data label or using the output of the intermediate layer. But, in the representation learning, the model learns the latent feature so, the label-based distillation process is not directly applicable to the representation learning. Also, for the case of fair representation distillation, we need to balance the distillation process in a way that the distillation performance should be maximized for both fairness and data utility.

\subsection{Synthetic Data Generations}


Generative models are used to create synthetic samples that has the same statistical properties of the original data. Most of the recent work in generative models are based on Variational AutoEncoder (VAE), Generative Adversarial Networks (GANs) and Diffusion model \cite{oussidi2018deep}. Variational AutoEncoder learns the data distribution by compressing the input data into lower dimensional latent space and reconstructing it back to the data space by using two architectures called Encoder and Decoder. Generative Adversarial Networks use adversarial techniques to learn data distribution by using noise as input in an architecture called generator and create some fake samples, while another architecture called Discriminator tries to distinguish between the real and fake samples and over time generator learns to fool the discriminator. In the diffusion model, noise is being added gradually to the input data itself until the data becomes pure noise then a neural network architecture is used to remove the noise from the data and in that process the neural network learns the data distribution.


\begin{definition}[Fair Generative Models (FGMs)]
A generative model $\mathcal{G}$ is said to be FGM iff the synthetic data it produces gives fair outcome in the downstreaming task.
\end{definition}

An example of a \emph{downstreaming task} for FGMs is attaining the DP and/or EO properties.





\section{Fair Representation Distillation and Synthetic Data Generation}
\label{sec:fair-representation-distillation}


In this work, we present a novel synthetic data generation technique by learning and distilling fair representation. We take the biased data $D$ and change the distribution $p(D)$ to fair distribution $p(D^{\prime})$. The process involves training fair representation, which minimizes the correlation between the sensitive and non-sensitive attributes, then distilling the fair representation and finally reconstruction of the high-fidelity data. We divide the whole process of the distillation and data generation into three stages:

\begin{enumerate}
	\item We take the biased data $D \in p(D)$ and encoded it into a fair latent space using variational auto encoder, in this process, the correlation between the sensitive and non-sensitive attributes is being minimized.
	\item We distill the fair latent space into a smaller model, this transfer the knowledge of learned fair representation from the VAE to the smaller model.
	\item We reconstruct high-fidelity fair synthetic data $D^{\prime} \in p(D^{\prime})$ using the distilled fair latent space and trained decoder from stage 1.  
\end{enumerate}


%


\subsection{Fair Representation Learning}

 Over the years, different approaches to learn the fair representation learning has been proposed and majority of them are based on the adversarial learning or using some sort of fairness constraints, e.g. demographic parity, equalized odds etc. \cite{zemel2013learning,zhang2018mitigating,ramachandranpillai2023fair}. In these works, fair representation is being learned by encoding the data into latent space $Z$ in a way that the correlation between the data features and sensitive attribute becomes lower. So, while doing a classification tasks, the outcome will be free from the sensitive attribute influence. To minimize the correlation between the sensitive attributes and other non-sensitive features, and to optimize the model for fair representation learning, mutual-information minimization \cite{rodriguez2021variational, nan2020variational, moyer2018invariant} and distance correlation minimization techniques \cite{liu2022fair, guo2022learning} has been proposed.  
 
 In this work, we use the distance correlation minimization approach equipped with variational autoencoder to learn the fair representation \cite{liu2022fair}. Given the dataset $(\{x_i, y_i, s_i\})_{i=1}^n \in D$, first we encode all the features using the encoder $\mathcal{E}_{\phi}$, into a low dimensional latent space $z \in \mathcal{Z}$, where, $z = \mathcal{E}_{\phi}(x, s)$. Here, $x \in \mathcal{X}$ and $s \in \mathcal{S}$ are the non-sensitive and sensitive attributes respectively. Then, we train the decoder ($\mathcal{D}_{\theta}$), and reconstruct the data $x^\prime = \mathcal{D}_{\theta}(z, s)$. Here, $\phi$ and $\theta$ are the parameters of the encoder and decoder respectively. To produce fair representation, we minimize the correlation between the latent space ($z$) and the sensitive attribute ($s$) using

\begin{equation}
	\label{eq:dis-correlation-min}
    \mathcal{V}^2_{\phi} (z, s) = \int_{\mathcal{Z}}\int_{\mathcal{S}} \mid p_{\phi}(z,s) - p_{\phi}(z) p(s) \mid^{2}\,dz\ \,ds.\
\end{equation}


The final optimization problem of fair representation learning is denoted by \cite{liu2022fair}:

\begin{equation}
    \argmax_{\phi, \theta}\{ \log p_{\theta}(x|s)-\beta \mathcal V_{\phi}^{2}(z,s)\}
\end{equation}

Here, $\mathcal{V}^2_{\phi} (z, s)$ is the distance covariance between the latent space $z$ and sensitive attribute $s$ \cite{liu2022fair}, and $\beta \in \{1,2,3,\dots,10\}$ is the hyper-parameter for balancing the level of fairness. 




\subsection{Distillation of Fair Latent Space}
\label{subsec:distillation-fair}

The main contribution of our work is to enabling the opportunity to distill the fair latent space. Most of the knowledge distillation works need data labels to transfer the knowledge to another model which is the reason they are not useful to learn representation learning.  Here we present a distribution matching distillation process where we use the trained encoder $\mathcal{E}_{\phi}$ to distill its knowledge to a smaller model $\mathcal{E}^{\prime}_{\psi}$. $\psi$ is the trainable parameter for the smaller model $\mathcal{E}^{\prime}$.
During the distillation process, the smaller model $\mathcal{E}^{\prime}_{\psi}$ takes the biased data $D$, as input and produce biased latent space $z^{\prime} = \mathcal{E}^{\prime}_{\psi}(x,s)$.
Then we transform the biased latent space by measuring the difference between the fair latent space and the biased latent space using a loss function $\mathcal{L}_\text{distillation}$, i.e.
\begin{equation}
\label{eq:l-distillation-loss}
	\mathcal{L}_{\text{distillation}}(z, z^\prime) = \sum_{j=1}^{k} \mathcal{L}_{\text{distillation}}(z_j, z_j^{\prime}).
\end{equation} 

Here, the choice of $\mathcal{L}_{\text{distillation}}$ can be \textit{L1 loss, MSE loss, mean-difference, Huber loss}. Besides ensuring fairness, to enforce the data utility features in the latent space, which helps to maximise the accuracy in the downstream task, we employ utility loss along the quality loss. We calculate the utility loss by calculating the KL divergence as follows:

\begin{equation}
\label{eq:kl}
	\mathcal{L}_\text{KL} = D_\text{KL}(q(z^{\prime}) \mid \mid p(z^{\prime}))
\end{equation} 


Here, $q(.)$ is an approximate model. So, the overall objective of our distillation process is

\begin{equation}
	\argmin_{\psi} {V(\mathcal{E}^{\prime}_{\psi})} = \underbrace{\mathcal{L}_{\text{distillation}}}_{\text{quality loss}} +  \underbrace{\lambda \times \mathcal{L}_\text{KL}}_{\text{utility loss}}
\end{equation}

Here, $\lambda$ is a tuning parameter to tweak the utility performance. In the experiment, we use $\lambda \in \{1,2,3,\dots,10\}$. Algorithm \ref{alg:algorithm} gives the overall process for fair distillation.



\subsection{Fair Synthetic Data Generation}

After successfully distilled the fair latent space in $\mathcal{E}^{\prime}_{\psi}$, we can use the distilled fair latent space and trained decoder $\mathcal{D}_\theta$ to reconstruct synthetic fair data, $x_{fair} \sim \mathcal{D}_{\theta}(z^{\prime})$, where $z^{\prime} \sim \mathcal{E}^{\prime}_{\psi}(x,s)$. Due to usage of quality loss ($\mathcal{L}_{\text{distillation}}$) and utility loss ($\mathcal{L}_\text{KL}$), the generated synthetic data have increase in both fairness and data utility in the downstreaming task. We show the result in section \ref{sec:experiments}.

\begin{algorithm}[t]
	\caption{Fair Representation Distillation Process}
	\label{alg:algorithm}
	\textbf{Input}: Biased dataset $D$, Trained Fair Encoder $\mathcal{E}_{\phi}$, Tuning parameter $\lambda$\\
	\textbf{Output}: Distilled model $\mathcal{E}^{\prime}_{\psi}$ \\
        \textbf{Initialize}: $\mathcal{E}^{\prime}_{\psi}$ randomly
	\begin{algorithmic}[1] 
		\FOR{each batch $(x_i, s_i) \sim D$}
		\STATE Sample $z \sim \mathcal{E}_{{\phi}}(x_i, s_i)$
            \STATE Sample $z^{\prime} \sim \mathcal{E}^{\prime}_\psi (x_i, s_i)$
            \STATE Calculate quality loss $\mathcal{L}_\text{distillation}$ using Eq.~\ref{eq:l-distillation-loss}
            \STATE Calculate data-utility loss $\mathcal{L}_\text{KL}$ using Eq.~\ref{eq:kl}
            \STATE $\mathcal{L}_\text{overall} \leftarrow \mathcal{L}_\text{distillation} + \lambda \times \mathcal{L}_\text{KL}$
            \STATE Update $\mathcal{E}^{\prime}_\psi$ using gradient descent w.r.t. $\mathcal{L}_\text{overall}$
		\ENDFOR
	\end{algorithmic}
\end{algorithm}

\section{Experiments}
\label{sec:experiments}

In this section, we present the experimental setup of our work. We use four benchmarking datasets (two tabular and two image) for this experiment and a wide range of evaluation metrics to verify the performance of the model. We compare the performance of our model with several state-of-the-art methods for the tabular data, e.g. Decaf, TabFairGAN, FairDisco, FLDGMs, pre-processing technique Correlation Remover and post-processing technique Threshold Optimizer. For the CelebA \cite{liu2015faceattributes} and Color MNIST \cite{lee2021learning} datasets, we perform the visualization analysis as well as the utility and fairness evaluation of our model. While training the benchmark, we use the same hyperparameters as in their respective original publication. For loading the dataset, training and comparing work with other state-of-the-art models, we use a fairness benchmarking tool called FairX \cite{sikder2024fairx}.

\subsection{Datasets}
\label{subsec:datasets}

In our experiment, we use two benchmarking tabular datasets (Adult income, COMPAS \cite{compascase}) and two image dataset (CelebA, Color MNIST), which are widely used in the data fairness field. Details of the dataset can be found in the supplementary section 1.

\subsection{Evaluation Metrics}
\label{subsec:evaluation-metrics}

We evaluate the performance of our model with the following criteria:

\begin{itemize}
    \item \textbf{Fairness:} We evaluate the fairness performance of our model by using the ``Demographic Parity Ratio (DPR)'' and ``Equalized Odds Ratio (EOR)'' in a downstream task (see definition \ref{def:dpr} and \ref{def:eor}). We use Random Forest as the classifier for the downstream task. The higher score in both ratios means high fairness.

    \item \textbf{Data Utility:} Beside fairness, we also evaluate the quality of our synthetic samples for data utility. Here, we also train a random forest classifier and measure the Accuracy, Recall and F1-score for the downstream task. 

    \item \textbf{Visual Evaluation:} To evaluate the distillation quality visually, we use ``t-SNE'' \cite{van2008visualizing} and ``PCA'' \cite{bryant1995principal} plots. These dimension reduction techniques, plot the latent space in two-dimensional space and show how closely the fair latent space and distilled fair latent space match each other.

    \item \textbf{Synthetic Data Quality:} We also evaluate the quality of the synthetic data. We check if the synthetic data is following the same distribution of the original data or not. We use ``Density'' and ``Coverage'' metrics \cite{naeem2020reliable, alaa2022faithful} to see how identical the two distributions are and how diverse the synthetic samples are respectively. The ``Coverage'' metrics also works as a detector of mode-dropping problem which is common issue for the generative models, especially for GANs \cite{goodfellow2020generative}. 

    \item \textbf{Explainability Analysis:} We show the feature importance on downstream tasks using an explainable algorithm \cite{lundberg2020local2global}. This helps us to understand the importance of each feature when making a decision.
    
\end{itemize}

        \begin{figure}
			 	\centering
			 	\includegraphics[width=0.85\columnwidth]{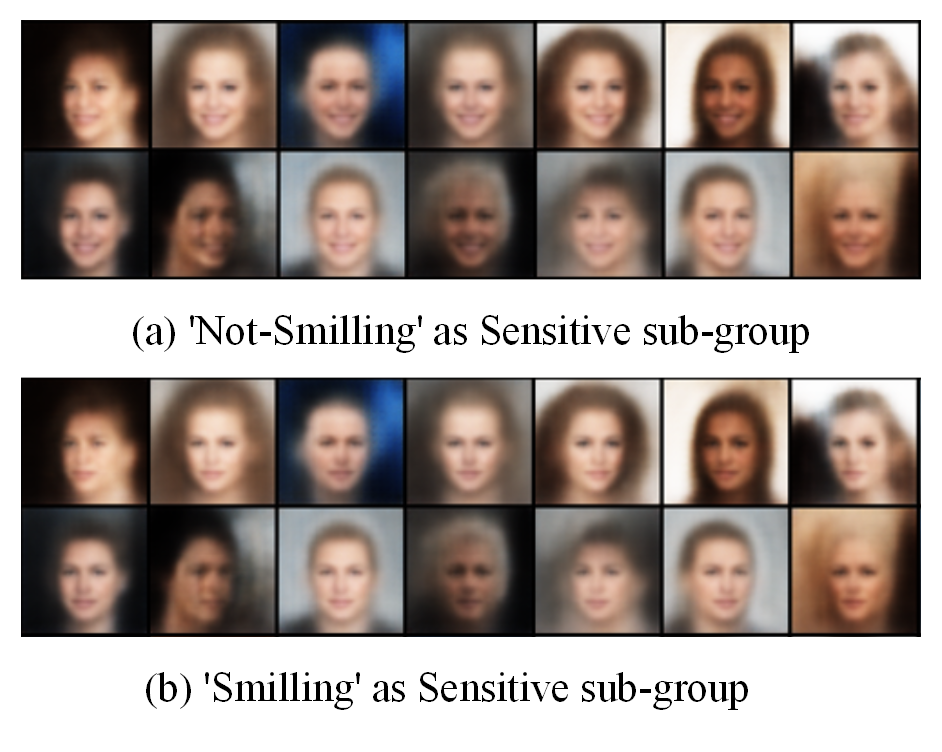}
			 	\caption{Generated Samples using Distilled fair representation, Dataset: CelebA, Sensitive\_attribute: Smiling}
			 	\label{fig:res-celeba-smilling}
        \end{figure}

\subsection{Experimental Setup}

In our experiment, for the model design, we follow the same model setup as presented in the FairDisco \cite{liu2022fair}. For the distilled encoder, we use the smaller model than the one presented in FairDisco. More about the hyperparameter settings can be found in the supplementary section 3. While training, we follow the splitting procedure of the dataset in 80-20 ratio for training and testing respectively presented by \citet{liu2022fair}. We run each evaluation metric ten times and report the mean and standard deviation.


\section{Results}
\label{sec:results}

     		\begin{table*}
			\caption{Evaluation on the \textbf{Adult-Income} dataset using different models including ours. Bold indicates best result, and all the metrics score are higher as better. Synthetic Utility is only applicable to
the Fair Generative Models (i.e. TabFairGAN, Decaf, FLDGMs and Our presented model).}
			\centering
				\resizebox{0.85\textwidth}{!}{%
					\begin{tabular}{lcccccccc}\toprule
						 & & \multicolumn{2}{c}{\textbf{Fairness Metrics}} &  \multicolumn{3}{c}{\textbf{Data Utility}} &  \multicolumn{2}{c}{\textbf{Synthetic Utility}}\\
						\cmidrule(lr){3-4}  \cmidrule(lr){5-7} \cmidrule(lr){8-9}
					   &\textbf{Protected} & DPR & EOR & ACC & Recall & F1- & Density & Coverage\\ 
						 & \textbf{Attribute} & & & & & Score & &\\

						\midrule
							
                       TabFairGAN & Gender  & 0.69 $\pm$ .01& 0.60 $\pm$ .01 & 0.84$\pm$ .01 & 0.61 $\pm$ .01 & 0.65 $\pm$ .01 & 0.006 $\pm$ .01 & 0.03$\pm$ .01\\
						& Race  & 0.026 $\pm$ .01 & 0.00 $\pm$ .00 & 0.84 $\pm$ .01 & 0.77 $\pm$ .01 & 0.67 $\pm$ .01 & 0.01 $\pm$ .01 & 0.02 $\pm$ .01\\
						\midrule
						
						Decaf  & Gender  & 0.52 $\pm$ .01& 0.42 $\pm$ .01 & 0.75 $\pm$ .01& 0.63 $\pm$ .01 & 0.44 $\pm$ .01 &  0.70 $\pm$ .01 & 0.571 $\pm$ .01\\
						 & Race & 0.55 $\pm$ .01 & 0.46 $\pm$ .01 & 0.77 $\pm$ .01 & 0.69 $\pm$ .01 & 0.53 $\pm$ .01 & 0.58 $\pm$ .01 & 0.84 $\pm$ .01\\
					
						\midrule

                         FLDGMs (DM) & Gender  &  0.94 $\pm$ .01 &  0.94 $\pm$ .01 &  0.69 $\pm$ .01 & 0.90 $\pm$ .01 & 0.81   $\pm$ .01  & 1.26 $\pm$ .01 & 0.89 $\pm$ .01\\
						 & Race  &  \textbf{0.99 $\pm$ .01} & 0.96 $\pm$ .01 & 0.69 $\pm$ .01 & 0.91 $\pm$ .01 & 0.81 $\pm$ .01  & 1.24 $\pm$ .01 & 0.86 $\pm$ .01\\
						\midrule
      
                             FairDisco (base-model)  & Gender  & 0.98 $\pm$ .01 & 0.85 $\pm$ .01 & 0.78 $\pm$ .01 & 0.92 $\pm$ .01 & 0.86 $\pm$ .01  & n/a & n/a \\
						 & Race  & 0.95 $\pm$ .01 & 0.92 $\pm$ .01 & 0.812 $\pm$ .01 & 0.71 $\pm$ .01 & 0.88 $\pm$ .01  & n/a & n/a \\
						\midrule
						
						Correlation-  & Gender  & 0.32 $\pm$ .01 & 0.23 $\pm$ .01 & 0.86 $\pm$ .01 & 0.65 $\pm$ .01 & 0.71 $\pm$ .01 & n/a & n/a  \\
						Remover & Race   & 0.29 $\pm$ .01 & 0.20 $\pm$ .01 & 0.86 $\pm$ .01 & 0.80 $\pm$ .01 & 0.71 $\pm$ .01  & n/a & n/a \\
						\midrule
						
						Threshold  & Gender   & 0.95 $\pm$ .01 & 0.35 $\pm$ .01 &0.86 $\pm$ .01 & 0.66 $\pm$ .01 & 0.65 $\pm$ .01 & n/a & n/a \\
						Optimizer& Race  & 0.69 $\pm$ .01 & 0.25 $\pm$ .01 & 0.87 $\pm$ .01 & 0.66 $\pm$ .01 & 0.71 $\pm$ .01 & n/a & n/a \\

                           \midrule

                           Original Data  & Gender  & 0.32 $\pm$ .01 & 0.22 $\pm$ .01& \textbf{0.88 $\pm$ .01 }& 0.80 $\pm$ .01& 0.72 $\pm$ .01 & n/a & n/a \\
						& Race  & 0.19 $\pm$ .01& 0.00 $\pm$ .00 &\textbf{0.88 $\pm$ .01}  & 0.80 $\pm$ .01& 0.71 $\pm$ .01 & n/a & n/a \\
                   
						\midrule
                           \midrule

                           Fair-distillation (ours)-L1-KL & Gender  &  \textbf{0.99 $\pm$ .01} &  \textbf{0.99 $\pm$ .01}& 0.76 $\pm$ .01 & \textbf{0.99 $\pm$ .01} &  \textbf{0.89 $\pm$ .01} & 1.06 $\pm$ .01 & \textbf{0.94 $\pm$ .01} \\
						& Race  & \textbf{0.99 $\pm$ .01} & \textbf{0.98 $\pm$ .01}& 0.76 $\pm$ .01 &\textbf{0.99  $\pm$ .01} & 0.87 $\pm$ .01 & 1.11 $\pm$ .01 & \textbf{0.97 $\pm$ .01} \\    

                            \midrule
                           Fair-distillation (ours)-Huber-KL & Gender  & 0.91 $\pm$ .01 & 0.84 $\pm$ .01& 0.82 $\pm$ .01  & 0.92 $\pm$ .01& 0.87 $\pm$ .01 & \textbf{1.96 $\pm$ .01} & 0.328 $\pm$ .01\\
						& Race  & 0.87 $\pm$ .01 & 0.84 $\pm$ .01& 0.84 $\pm$ .01 & 0.92  $\pm$ .01 & 0.89 $\pm$ .01 & 1.65 $\pm$ .01 & 0.652 $\pm$ .01\\

                          \midrule
                           Fair-distillation (ours)-MSE-KL & Gender  & 0.90 $\pm$ .01 & 0.87 $\pm$ .01& 0.82 $\pm$ .01  & 0.92 $\pm$ .01& \textbf{0.89 $\pm$ .01} & 1.86 $\pm$ .01 & 0.332 $\pm$ .01\\
						& Race  & 0.87 $\pm$ .01 & 0.78 $\pm$ .01& 0.84 $\pm$ .01 & 0.92  $\pm$ .01 & \textbf{0.90 $\pm$ .01} & \textbf{1.68 $\pm$ .01} & 0.67 $\pm$ .01\\

                          \midrule
                           Fair-distillation (ours)-MD-KL & Gender  & 0.98 $\pm$ .01 & 0.90 $\pm$ .01& 0.76 $\pm$ .01  & 0.93 $\pm$ .01& 0.86 $\pm$ .01 & 1.54 $\pm$ .01 & 0.89 $\pm$ .01\\
						& Race  & 0.94 $\pm$ .01 & \textbf{0.98 $\pm$ .01}& 0.78 $\pm$ .01 & 0.91  $\pm$ .01 & 0.86 $\pm$ .01 & 1.43 $\pm$ .01 & 0.90 $\pm$ .01\\
						
						\bottomrule
					\end{tabular}
				
				\label{tab:table1}
				}
		\end{table*}

In this section, we present our results and comparison with the state-of-the-art works. We compare our work with six bias-mitigation models (including fair generative models) and our presented work outperform them in the terms of fairness, data utility and synthetic sample quality. 

\subsection{Tabular Data Results} First, we present the performance of our architecture for the tabular data. Table \ref{tab:table1} shows the data utility and fairness result for the ``Adult-Income'' dataset.
\footnote{For the page limitation, the results of ``Compas'' dataset have been added in the supplementary section 2.1} 
We show the result of our experiment using two sensitive attributes \{sex, race\}. Also, for the downstream task, we predict the income class for the individuals. Besides the benchmarking model, we show the result using different variations of loss functions on our model (described in section \ref{subsec:distillation-fair}). We observe that L1-loss with KL divergence achieves best performance in terms of fairness, data utility and synthetic quality. Our distilled model achieves better performance than the base model as well as with other architectures. We get near-perfect results for both of the protected attributes in terms of fairness in both metrics (DPR \& EOR score 0.99 and 0.98). We achieve this with the help of our proposed quality and utility loss, which helps us gain better performance in fairness and utility than the fair base model and other competing models. We achieve a 5\% and 10\% rise in the fairness and utility performance compared to state-of-the-art (FLDGMs). We also show the effect of the tuning parameter $\lambda$ in supplementary Figure 1, where $\lambda = 1$ exhibits the best performance.

        \begin{figure}[t]
			 	\centering
			 	\includegraphics[width=\columnwidth]{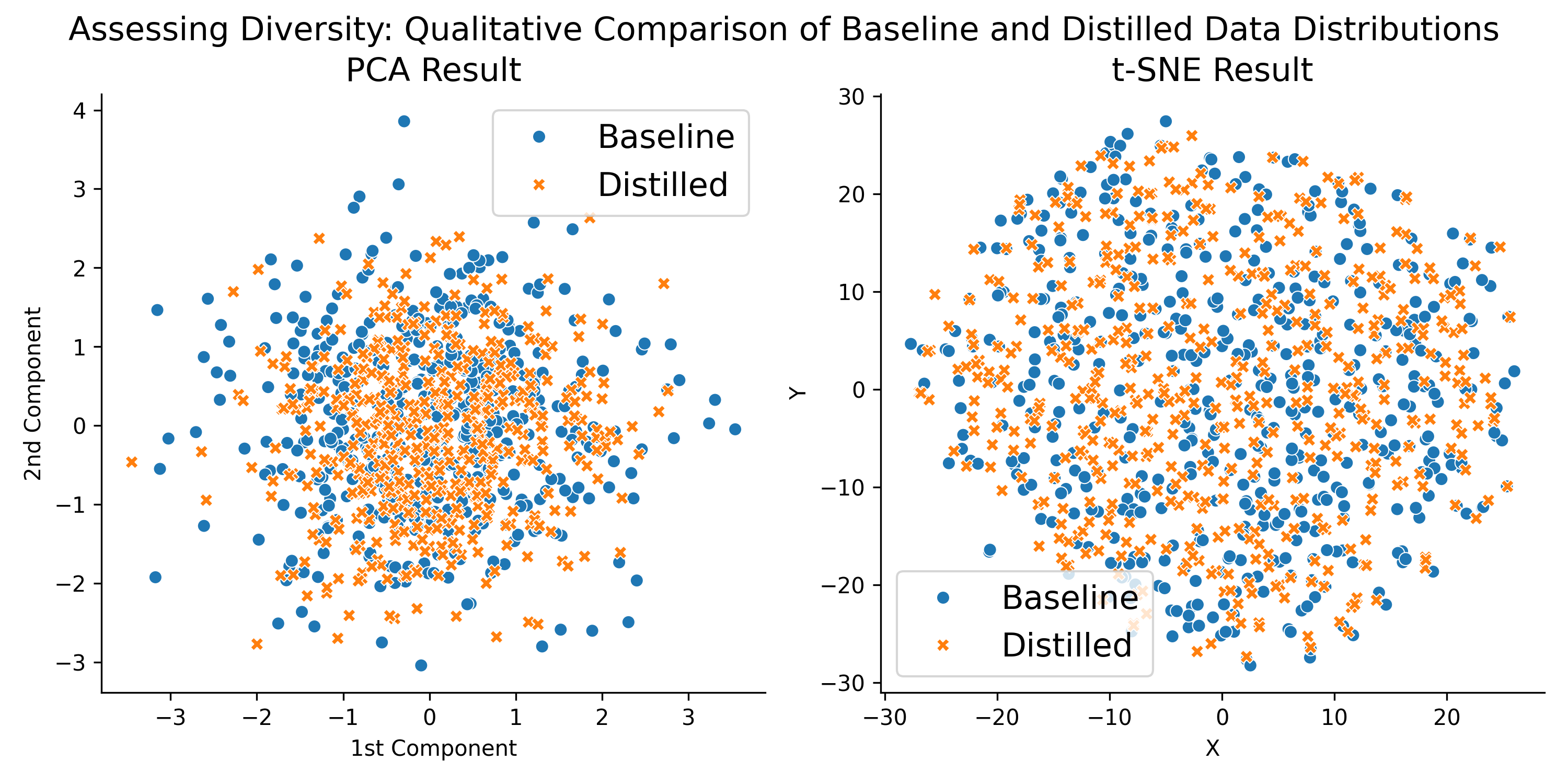}
			 	\caption{PCA (left) and t-SNE (right) plots of the fair distribution (orange circle) and distilled distribution (blue circle), here each dot represents each record in the dataset, if the distilled model learns the data distribution perfectly, then both colored dot should overlap with each other, Dataset: Compass}
			 	\label{fig:pca-tsne}
        \end{figure}

\subsection{Image Data Results} We train the CelebA dataset with the target ``Gender'' and sensitive attribute ``Smiling-status'', where ``Smilling'' and ``Not smilling'' are the sub-groups. Figure \ref{fig:res-celeba-smilling} shows the synthetic data generated by our model, where we use the \{Gender, Smiling\}, \{Gender, Not-Smiling\} combination. For the Color MNIST, we use the ``digit number'' as the target and ``colors'' as sensitive attributes. We use the \{Digit, Red\}, \{Digit, Blue\}, \{Digit, red\} combinations for this particular dataset \footnote{We show the result of Color-MNIST in the supplementary section 2.2, including the synthetic utility measurement of the images.}. We observe, our model generates images based on the sensitive sub-groups accurately.

\subsection{Visual Evaluation} To show if the smaller distilled model is, in fact, distilling the fair representation or not, we use PCA and t-SNE plots to project both latent space (fair and distilled fair) into two-dimensional space. Figure \ref{fig:pca-tsne} shows the PCA and t-SNE plots of the fair-representation and distilled representation for the ``Compas'' dataset, sensitive attribute: ``sex''. Each dot represents a data point. Here, we observe both data points from the fair representation and distilled representation almost overlap with each other; this means the distilled representation has captured the fair data distribution.

        \begin{figure}[t]
			 	\centering
			 	\includegraphics[width=0.8\columnwidth]{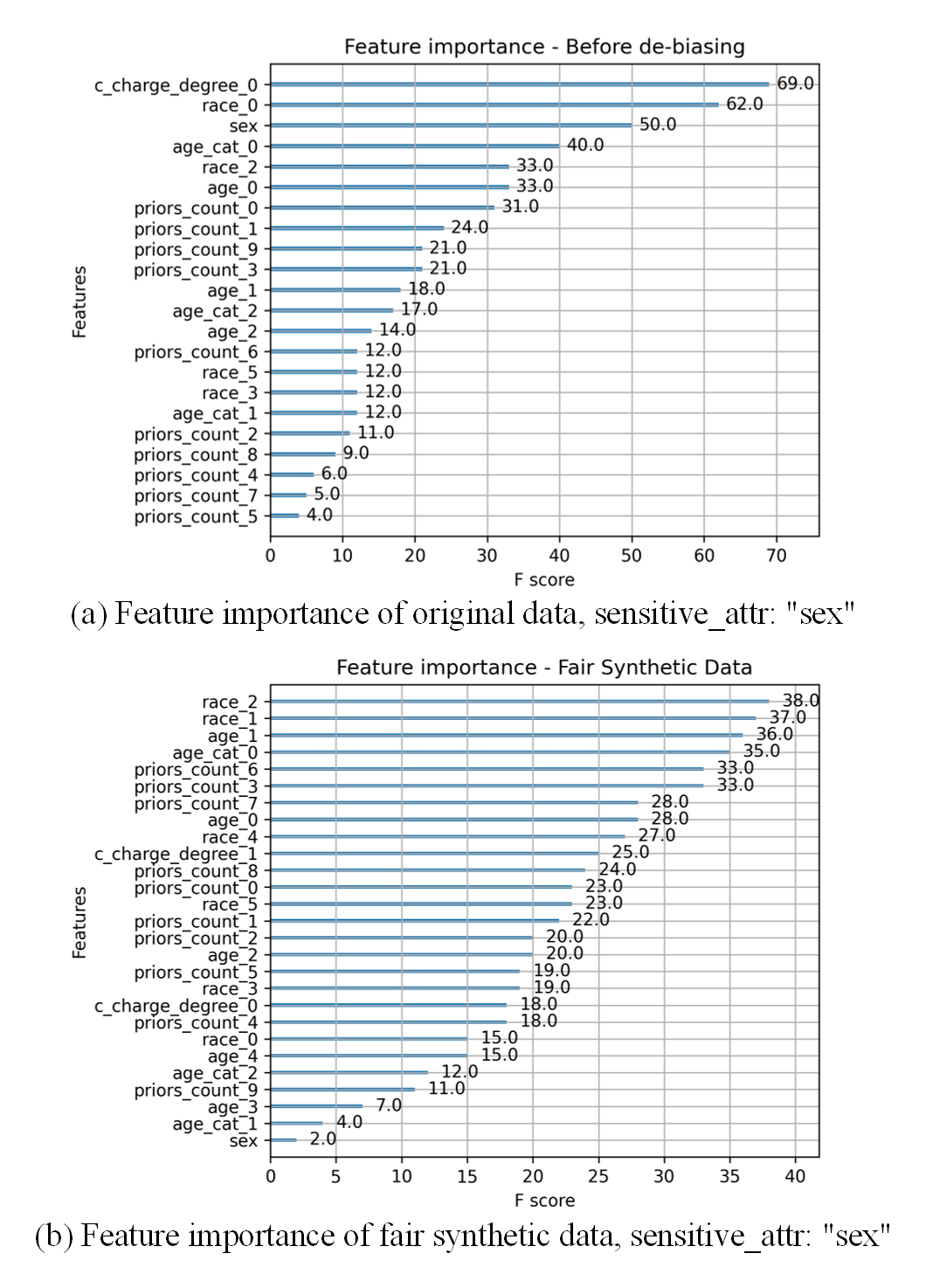}
			 	\caption{Explainability Analysis: (a) shows the feature importance of using the original data, where the sensitive attribute: ``sex'' is showing high feature importance, (b) shows the feature importance of our generated samples, where the sensitive attribute is showing low feature importance}
			 	\label{fig:feature-importance}
        \end{figure}

\subsection{Synthetic Data Analysis} Beside the fairness analysis and data utility analysis, we also perform the synthetic data quality analysis. Where we check the quality of the synthetic data. Generative model should follow the distribution of the original data and the generated samples should be diverse. We use the ``Density'' and ``Coverage'' metrics to check if the data generated by the generative model follows the original data distribution and if the synthetic samples are diverse. We observe from Table \ref{tab:table1}, that different variations of our model achieve better scores for both of the synthetic quality metrics, especially our ``L1-KL'' variation achieves better performance in all three criteria (fairness, utility and synthetic quality). So this signifies that our generative model produces high-quality and more diverse samples than other generative models. 


\subsection{Explainability Analysis} For this analysis, we take the fair synthetic data generated from our model, then train a tree-based classifier for a downstream task, and show the feature importance of both biased data and synthetic data. We use the ``Compas'' dataset for this task and we predict the recidivism rates of the defendants in two years. Figure \ref{fig:feature-importance} shows the explainability analysis, where the sensitive attribute is ``Sex (Gender)''. We observe, while doing the downstream task on the biased dataset, the importance of the sensitive feature is too high (F Score) in the prediction task. On the contrary, when using our generated synthetic data, the feature importance of the sensitive feature is negligible. This signifies the fair quality of our generated data.  

\subsection{Run-time Analysis} 
\label{subsec:runtime}

For the run time analysis of different generative model in our study (TabFairGAN, FLDGMs, Decaf and ours), we measure the training time for the models. We use the same machine and dataset to train these models and record the time. Our experiment shows, our model use 23\% less time than the FLDGMs, 105\% less than TabFairGAN and 153\% less time than Decaf. This happen due to the usage of simple architecture than the other models also for training in the latent space. \footnote{More detailing result of the run-time analysis can be found in the supplementary section 2.3.}


\subsection{Discussion}
\label{subsec:discussion}

To improve the fairness quality, data quality in the synthetic data also reduce the computational overhead in the generative model, in this work, we present a distillation based fair generative model. We present our approach in the section \ref{sec:fair-representation-distillation}, where we use quality and utility loss to distill the latent space. We use a small model to distill the both tabular and image datasets, this aids to reduce the computational overhead while training the model and take less time (analysis in section \ref{subsec:runtime}). We also show in the Table \ref{tab:table1} that our model is performing better than the distillation base model (FairDisco) as well as all other benchmarking models present in this study. We use wide range of evaluation metrics to evaluate the quality of our generated samples, in terms of fairness, data utility, synthetic data quality, explainability analysis and visual evaluation. We show the distilled model is able to capture the data distribution of the fair base model (Figure \ref{fig:pca-tsne} and ``Density'' metrics in Table \ref{tab:table1}). We also show the synthetic samples of Image dataset in the Figure \ref{fig:res-celeba-smilling}.

\paragraph{Limitations \& Future Works} 

Amidst doing distillation in the feature space, reducing computational cost and generating high-quality fair data, while creating the fair representation, we are considering single protected attribute. In the future work, we will focus on extending the work for multiple sensitive attribute. However, it is noteworthy that, as we are distilling from the latent space, so the same architecture and approach we present here can be used to distill the latent space for multiple sensitive attributes.

\paragraph{Social Impact} 

As a fair generative model, our work takes the biased data and creates both fair representation and fair synthetic data. Using biased data while decision-making might make the decision biased towards some demographics, e.g. the Dutch scandal for fraud detection we discuss in the section \ref{sec:introduction} \cite{dutchscandal}. Synthetic fair data and fair representation can mitigate the gap, as these synthetic samples and representations do not take sensitive attributes into consideration while making any decision.

\section{Conclusion}
\label{sec:conclusion}


With the advancement of AI-enabled decision making system, the necessity of fair model has increased due to their vulnerability towards bias decision. Over the times, researchers present various bias-mitigation techniques including generative models to create fair representation and/or fair synthetic samples. However, these generative models need extensive computational resources. One way to mitigate this issue is to use knowledge distillation to transfer the knowledge from one model to another smaller model. But existing distillation methods works on labelled-dataset which makes it challenging to distill representation. From these motivation, we present a novel fair generative model, where we present an approach to distill fair latent space and use that to generate high-fidelity fair synthetic samples on both tabular and image dataset. As we are distilling in the latent space and smaller model, we require less computational resources. We test our model with four widely used dataset (tabular, image) and in terms of fairness, utility and synthetic utility and our model outperforms the state-of-the-art models. We believe the model can help to mitigate the unfair decision making issue in the society.


%

\bibliography{aaai25}

\begin{thebibliography}{39}
\providecommand{\natexlab}[1]{#1}

\bibitem[{Alaa et~al.(2022)Alaa, Van~Breugel, Saveliev, and van~der
  Schaar}]{alaa2022faithful}
Alaa, A.; Van~Breugel, B.; Saveliev, E.~S.; and van~der Schaar, M. 2022.
\newblock How faithful is your synthetic data? sample-level metrics for
  evaluating and auditing generative models.
\newblock In \emph{International Conference on Machine Learning}, 290--306.
  PMLR.

\bibitem[{Angwin et~al.(2016)Angwin, Larson, Mattu, and Kirchner}]{compascase}
Angwin, J.; Larson, J.; Mattu, S.; and Kirchner, L. 2016.
\newblock Machine Bias: There’s software used across the country to predict
  future criminals. And it’s biased against blacks.
\newblock
  \url{https://www.propublica.org/article/machine-bias-risk-assessments-in-criminal-sentencing}.
\newblock [Online; accessed 12-Aug-2024].

\bibitem[{Barocas and Selbst(2016)}]{barocas2016big}
Barocas, S.; and Selbst, A.~D. 2016.
\newblock Big data's disparate impact.
\newblock \emph{Calif. L. Rev.}, 104: 671.

\bibitem[{Bryant and Yarnold(1995)}]{bryant1995principal}
Bryant, F.~B.; and Yarnold, P.~R. 1995.
\newblock {Principal-Components Analysis and Exploratory and Confirmatory
  Factor Analysis}.

\bibitem[{Caton and Haas(2024)}]{caton2024fairness}
Caton, S.; and Haas, C. 2024.
\newblock Fairness in machine learning: A survey.
\newblock \emph{ACM Computing Surveys}, 56(7): 1--38.

\bibitem[{Dai et~al.(2021)Dai, Jiang, Wu, Bao, Wang, Liu, and
  Zhou}]{dai2021general}
Dai, X.; Jiang, Z.; Wu, Z.; Bao, Y.; Wang, Z.; Liu, S.; and Zhou, E. 2021.
\newblock General instance distillation for object detection.
\newblock In \emph{Proceedings of the IEEE/CVF conference on computer vision
  and pattern recognition}, 7842--7851.

\bibitem[{Dong et~al.(2023)Dong, Zhang, Yuan, Zou, Wang, and
  Li}]{dong2023reliant}
Dong, Y.; Zhang, B.; Yuan, Y.; Zou, N.; Wang, Q.; and Li, J. 2023.
\newblock Reliant: Fair knowledge distillation for graph neural networks.
\newblock In \emph{Proceedings of the 2023 SIAM International Conference on
  Data Mining (SDM)}, 154--162. SIAM.

\bibitem[{Dutta et~al.(2020)Dutta, Wei, Yueksel, Chen, Liu, and
  Varshney}]{dutta2020there}
Dutta, S.; Wei, D.; Yueksel, H.; Chen, P.-Y.; Liu, S.; and Varshney, K. 2020.
\newblock Is there a trade-off between fairness and accuracy? a perspective
  using mismatched hypothesis testing.
\newblock In \emph{International conference on machine learning}, 2803--2813.
  PMLR.

\bibitem[{Fang et~al.(2024)Fang, Che, Mao, Zhang, Zhao, and
  Zhao}]{fang2024bias}
Fang, X.; Che, S.; Mao, M.; Zhang, H.; Zhao, M.; and Zhao, X. 2024.
\newblock Bias of AI-generated content: an examination of news produced by
  large language models.
\newblock \emph{Scientific Reports}, 14(1): 5224.

\bibitem[{Gao et~al.(2022)Gao, Zhai, Ma, Shen, Chen, and
  Wang}]{gao2022fairneuron}
Gao, X.; Zhai, J.; Ma, S.; Shen, C.; Chen, Y.; and Wang, Q. 2022.
\newblock FairNeuron: improving deep neural network fairness with adversary
  games on selective neurons.
\newblock In \emph{Proceedings of the 44th International Conference on Software
  Engineering}, 921--933.

\bibitem[{Goodfellow et~al.(2020)Goodfellow, Pouget-Abadie, Mirza, Xu,
  Warde-Farley, Ozair, Courville, and Bengio}]{goodfellow2020generative}
Goodfellow, I.; Pouget-Abadie, J.; Mirza, M.; Xu, B.; Warde-Farley, D.; Ozair,
  S.; Courville, A.; and Bengio, Y. 2020.
\newblock Generative adversarial networks.
\newblock \emph{Communications of the ACM}, 63(11): 139--144.

\bibitem[{Guo et~al.(2022)Guo, Wang, Wang, and Zha}]{guo2022learning}
Guo, D.; Wang, C.; Wang, B.; and Zha, H. 2022.
\newblock Learning fair representations via distance correlation minimization.
\newblock \emph{IEEE Transactions on Neural Networks and Learning Systems},
  35(2): 2139--2152.

\bibitem[{Heikkil\"a(2021)}]{dutchscandal}
Heikkil\"a, M. 2021.
\newblock Dutch scandal serves as a warning for Europe over risks of using
  algorithms.
\newblock
  \url{https://www.politico.eu/article/dutch-scandal-serves-as-a-warning-for-europe-over-risks-of-using-algorithms/}.
\newblock [Online; accessed 12-Aug-2024].

\bibitem[{Hinton, Vinyals, and Dean(2015)}]{hinton2015distilling}
Hinton, G.; Vinyals, O.; and Dean, J. 2015.
\newblock Distilling the Knowledge in a Neural Network.
\newblock \emph{stat}, 1050: 9.

\bibitem[{Jung et~al.(2021)Jung, Lee, Park, and Moon}]{jung2021fair}
Jung, S.; Lee, D.; Park, T.; and Moon, T. 2021.
\newblock Fair feature distillation for visual recognition.
\newblock In \emph{Proceedings of the IEEE/CVF conference on computer vision
  and pattern recognition}, 12115--12124.

\bibitem[{Knobbout(2023)}]{maxalfr2023}
Knobbout, M. 2023.
\newblock {ALFR++: A Novel Algorithm for Learning Adversarial Fair
  Representations}.
\newblock In \emph{ECAI 2023}, 1280--1287. IOS Press.

\bibitem[{Lee et~al.(2021)Lee, Kim, Lee, Lee, and Choo}]{lee2021learning}
Lee, J.; Kim, E.; Lee, J.; Lee, J.; and Choo, J. 2021.
\newblock Learning debiased representation via disentangled feature
  augmentation.
\newblock \emph{Advances in Neural Information Processing Systems}, 34:
  25123--25133.

\bibitem[{Li, Ren, and Deng(2022)}]{li2022fairgan}
Li, J.; Ren, Y.; and Deng, K. 2022.
\newblock Fairgan: Gans-based fairness-aware learning for recommendations with
  implicit feedback.
\newblock In \emph{Proceedings of the ACM web conference 2022}, 297--307.

\bibitem[{Liu et~al.(2022)Liu, Li, Yao, Xu, Ma, Xu, and Tong}]{liu2022fair}
Liu, J.; Li, Z.; Yao, Y.; Xu, F.; Ma, X.; Xu, M.; and Tong, H. 2022.
\newblock Fair representation learning: An alternative to mutual information.
\newblock In \emph{Proceedings of the 28th ACM SIGKDD Conference on Knowledge
  Discovery and Data Mining}, 1088--1097.

\bibitem[{Liu et~al.(2015)Liu, Luo, Wang, and Tang}]{liu2015faceattributes}
Liu, Z.; Luo, P.; Wang, X.; and Tang, X. 2015.
\newblock Deep Learning Face Attributes in the Wild.
\newblock In \emph{Proceedings of International Conference on Computer Vision
  (ICCV)}.

\bibitem[{Lundberg et~al.(2020)Lundberg, Erion, Chen, DeGrave, Prutkin, Nair,
  Katz, Himmelfarb, Bansal, and Lee}]{lundberg2020local2global}
Lundberg, S.~M.; Erion, G.; Chen, H.; DeGrave, A.; Prutkin, J.~M.; Nair, B.;
  Katz, R.; Himmelfarb, J.; Bansal, N.; and Lee, S.-I. 2020.
\newblock From local explanations to global understanding with explainable AI
  for trees.
\newblock \emph{Nature Machine Intelligence}, 2(1): 2522--5839.

\bibitem[{Madras et~al.(2018)Madras, Creager, Pitassi, and
  Zemel}]{madras2018learning}
Madras, D.; Creager, E.; Pitassi, T.; and Zemel, R. 2018.
\newblock Learning adversarially fair and transferable representations.
\newblock In \emph{International Conference on Machine Learning}, 3384--3393.
  PMLR.

\bibitem[{Moyer et~al.(2018)Moyer, Gao, Brekelmans, Galstyan, and
  Ver~Steeg}]{moyer2018invariant}
Moyer, D.; Gao, S.; Brekelmans, R.; Galstyan, A.; and Ver~Steeg, G. 2018.
\newblock Invariant representations without adversarial training.
\newblock \emph{Advances in neural information processing systems}, 31.

\bibitem[{Naeem et~al.(2020)Naeem, Oh, Uh, Choi, and Yoo}]{naeem2020reliable}
Naeem, M.~F.; Oh, S.~J.; Uh, Y.; Choi, Y.; and Yoo, J. 2020.
\newblock Reliable fidelity and diversity metrics for generative models.
\newblock In \emph{International Conference on Machine Learning}, 7176--7185.
  PMLR.

\bibitem[{Nan and Tao(2020)}]{nan2020variational}
Nan, L.; and Tao, D. 2020.
\newblock Variational approach for privacy funnel optimization on continuous
  data.
\newblock \emph{Journal of Parallel and Distributed Computing}, 137: 17--25.

\bibitem[{Oussidi and Elhassouny(2018)}]{oussidi2018deep}
Oussidi, A.; and Elhassouny, A. 2018.
\newblock Deep generative models: Survey.
\newblock In \emph{2018 International conference on intelligent systems and
  computer vision (ISCV)}, 1--8. IEEE.

\bibitem[{Rajabi and Garibay(2022)}]{rajabi2022tabfairgan}
Rajabi, A.; and Garibay, O.~O. 2022.
\newblock Tabfairgan: Fair tabular data generation with generative adversarial
  networks.
\newblock \emph{Machine Learning and Knowledge Extraction}, 4(2): 488--501.

\bibitem[{Ramachandranpillai, Sikder, and
  Heintz(2023)}]{ramachandranpillai2023fair}
Ramachandranpillai, R.; Sikder, M.~F.; and Heintz, F. 2023.
\newblock {Fair Latent Deep Generative Models (FLDGMs) for Syntax-Agnostic and
  Fair Synthetic Data Generation}.
\newblock In \emph{ECAI 2023}, 1938--1945. IOS Press.

\bibitem[{Rodr{\'\i}guez-G{\'a}lvez, Thobaben, and
  Skoglund(2021)}]{rodriguez2021variational}
Rodr{\'\i}guez-G{\'a}lvez, B.; Thobaben, R.; and Skoglund, M. 2021.
\newblock A variational approach to privacy and fairness.
\newblock In \emph{2021 IEEE Information Theory Workshop (ITW)}, 1--6. IEEE.

\bibitem[{Sikder et~al.(2024)Sikder, Ramachandranpillai, de~Leng, and
  Heintz}]{sikder2024fairx}
Sikder, M.~F.; Ramachandranpillai, R.; de~Leng, D.; and Heintz, F. 2024.
\newblock FairX: A comprehensive benchmarking tool for model analysis using
  fairness, utility, and explainability.
\newblock \emph{arXiv preprint arXiv:2406.14281}.

\bibitem[{Van~Breugel et~al.(2021)Van~Breugel, Kyono, Berrevoets, and Van~der
  Schaar}]{van2021decaf}
Van~Breugel, B.; Kyono, T.; Berrevoets, J.; and Van~der Schaar, M. 2021.
\newblock Decaf: Generating fair synthetic data using causally-aware generative
  networks.
\newblock \emph{Advances in Neural Information Processing Systems}, 34:
  22221--22233.

\bibitem[{Van~der Maaten and Hinton(2008)}]{van2008visualizing}
Van~der Maaten, L.; and Hinton, G. 2008.
\newblock {Visualizing Data using T-SNE}.
\newblock \emph{Journal of machine learning research}, 9(11).

\bibitem[{Weerts et~al.(2023)Weerts, Dud{\'\i}k, Edgar, Jalali, Lutz, and
  Madaio}]{weerts2023fairlearn}
Weerts, H.; Dud{\'\i}k, M.; Edgar, R.; Jalali, A.; Lutz, R.; and Madaio, M.
  2023.
\newblock Fairlearn: Assessing and Improving Fairness of AI Systems.
\newblock \emph{Journal of Machine Learning Research}, 24(257): 1--8.

\bibitem[{Weerts, Theunissen, and Willemsen(2023)}]{weerts2023look}
Weerts, H.~J.; Theunissen, R.; and Willemsen, M.~C. 2023.
\newblock Look and You Will Find It: Fairness-Aware Data Collection through
  Active Learning.
\newblock In \emph{IAL@ PKDD/ECML}, 74--88.

\bibitem[{Xu et~al.(2021)Xu, Liu, Li, Jain, and Tang}]{xu2021robust}
Xu, H.; Liu, X.; Li, Y.; Jain, A.; and Tang, J. 2021.
\newblock To be robust or to be fair: Towards fairness in adversarial training.
\newblock In \emph{International conference on machine learning}, 11492--11501.
  PMLR.

\bibitem[{Zemel et~al.(2013)Zemel, Wu, Swersky, Pitassi, and
  Dwork}]{zemel2013learning}
Zemel, R.; Wu, Y.; Swersky, K.; Pitassi, T.; and Dwork, C. 2013.
\newblock Learning fair representations.
\newblock In \emph{International conference on machine learning}, 325--333.
  PMLR.

\bibitem[{Zhang, Lemoine, and Mitchell(2018)}]{zhang2018mitigating}
Zhang, B.~H.; Lemoine, B.; and Mitchell, M. 2018.
\newblock Mitigating unwanted biases with adversarial learning.
\newblock In \emph{Proceedings of the 2018 AAAI/ACM Conference on AI, Ethics,
  and Society}, 335--340.

\bibitem[{Zhang and Ma(2020)}]{zhang2020improve}
Zhang, L.; and Ma, K. 2020.
\newblock Improve object detection with feature-based knowledge distillation:
  Towards accurate and efficient detectors.
\newblock In \emph{International Conference on Learning Representations}.

\bibitem[{Zhu et~al.(2024)Zhu, Li, Chen, and Zheng}]{zhu2024devil}
Zhu, Y.; Li, J.; Chen, L.; and Zheng, Z. 2024.
\newblock The Devil is in the Data: Learning Fair Graph Neural Networks via
  Partial Knowledge Distillation.
\newblock In \emph{Proceedings of the 17th ACM International Conference on Web
  Search and Data Mining}, 1012--1021.

\end{thebibliography}

\end{document}